\pdfoutput=1
\documentclass[11pt]{article}
\usepackage{graphicx}
\graphicspath{ {./diagrams/} }
\usepackage{authblk}

\usepackage{acl}

\usepackage{times}
\usepackage{graphicx}
\usepackage{latexsym}
\usepackage{amsmath}
\usepackage{amssymb}
\usepackage{booktabs}
\usepackage[T1]{fontenc}

\usepackage[utf8]{inputenc}

\usepackage[symbol]{footmisc}

\usepackage{microtype}

\usepackage{inconsolata}

\usepackage[fitting]{tcolorbox}
\usepackage{lipsum}

\newtcboxfit{\mybox}[1][]{%
  fit algorithm=hybrid*,
  width=3in,
  height=5in,
  sharp corners,
  colframe=black,colback=white,
  size=fbox,#1
}

\title{Low-Cost Generation and Evaluation of Dictionary Example Sentences}

\makeatletter
\newcommand\email[2][]%
   {\newaffiltrue\let\AB@blk@and\AB@pand
      \if\relax#1\relax\def\AB@note{\AB@thenote}\else\def\AB@note{\relax}%
        \setcounter{Maxaffil}{0}\fi
      \begingroup
        \let\protect\@unexpandable@protect
        \def\thanks{\protect\thanks}\def\footnote{\protect\footnote}%
        \@temptokena=\expandafter{\AB@authors}%
        {\def\\{\protect\\\protect\Affilfont}\xdef\AB@temp{#2}}%
         \xdef\AB@authors{\the\@temptokena\AB@las\AB@au@str
         \protect\\[\affilsep]\protect\Affilfont\AB@temp}%
         \gdef\AB@las{}\gdef\AB@au@str{}%
        {\def\\{, \ignorespaces}\xdef\AB@temp{#2}}%
        \@temptokena=\expandafter{\AB@affillist}%
        \xdef\AB@affillist{\the\@temptokena \AB@affilsep
          \AB@affilnote{}\protect\Affilfont\AB@temp}%
      \endgroup
       \let\AB@affilsep\AB@affilsepx
}
\makeatother

\author[1]{Bill Cai$\dag$}
\author[2]{Clarence Ng$\dag$}
\author[2]{Daniel Tan$\ddag$}
\author[1]{Shelvia Hotama$\ddag$}
\affil[1]{Amazon Web Services}
\affil[2]{Ministry of Education, Singapore}
\email{\url{{billcaiy,hshelvia}@amazon.com}}
\email{\url{{clarence_ng,daniel_tan}@moe.gov.sg}}

\begin{document}
\maketitle
\footnotetext{$\dag$, $\ddag$ Equal Contribution}
\setcounter{footnote}{0}
\begin{abstract}
Dictionary example sentences play an important role in illustrating word definitions and usage, but manually creating quality sentences is challenging. Prior works have demonstrated that language models can be trained to generate example sentences. However, they relied on costly customized models and word sense datasets for generation and evaluation of their work. Rapid advancements in foundational models present the opportunity to create low-cost, zero-shot methods for the generation and evaluation of dictionary example sentences. We introduce a new automatic evaluation metric called OxfordEval that measures the win-rate of generated sentences against existing Oxford Dictionary sentences. OxfordEval shows high alignment with human judgments, enabling large-scale automated quality evaluation. We experiment with various LLMs and configurations to generate dictionary sentences across word classes. We complement this with a novel approach of using masked language models to identify and select sentences that best exemplify word meaning. The eventual model, FM-MLM, achieves over 85.1\% win rate against Oxford baseline sentences according to OxfordEval, compared to 39.8\% win rate for prior model-generated sentences. 
\end{abstract}

\section{Introduction}

Dictionary example sentences play a vital role in illustrating the meanings and usage of headwords for dictionary users. Prior studies have found evidence supporting the importance of the quantity and quality of example sentences in improving learners' performance for receptive and productive language tasks \citep{nesi1996role,frankenberg2014use}. 

Creating and maintaining example sentences for dictionaries can be a labour-intensive task – the Oxford Dictionary of English \citep{stevenson2010oxford}, for example, aims to represent the current usage of close to 100,000 English headwords. Traditional efforts to improve productivity of example sentence creation focused on searching for sentences with headwords in existing electronic text corpus \citep{kilgarriff2008gdex,hanks2009impact,frankenberg2021slipping}. However, such text corpora consists of content for myriad purposes, and may not always contain example sentences that best exemplify the meanings and usage of words to support language learning. To address this, recent works have demonstrated that language models can be trained to generate new sentences on unseen headwords \citep{he2022controllable, harvill2023one}, vastly expanding the search space that can be considered for crafting quality sentences. However, they often rely on custom-trained models and datasets annotated with word senses for the generation and evaluation of sentences, which can be a costly process. 

Rapid advancements in foundational models (FMs) \cite{bommasani2021opportunities} now offer new possibilities for more flexible and creative generation of dictionary example sentences at low cost. Closed and open-source large language models (LLMs) like Claude and Llama-2 have demonstrated an impressive capacity to generate fluent, coherent text while capturing nuances of style, tone and topic \citep{liang2022holistic}. Their exposure to large corpora of linguistic data during pretraining \citep{touvron2023llama, penedo2023refinedweb} and ability to follow specific instructions \citep{wang2022super, mishra2022cross} allow these models to perform well for new previously unseen instructional tasks \citep{radford2019language}. They offer the potential for dictionary example sentences to be generated and evaluated at scale without the use of specialised models or datasets. 

This paper examines low-cost, zero-shot methods for LLMs to automatically generate and evaluate dictionary example sentences. We start by defining the OxfordEval metric, which measures the win-rate of candidate dictionary example sentences when evaluated competitively by LLMs for quality against samples from the Oxford Dictionary. Using human annotations on a subset of outputs, we validate that OxfordEval matches well with human preferences. With this validated automatic evaluation measure, we are then able to expansively test various state-of-the-art models including Claude, Llama-2 and Mistral, and analyse how variations in generation methodology affect sentence quality. We show that LLMs can generate sentences that are preferred to Oxford Dictionary example sentences 83.9\% of the time, while past model-generated sentences only have a win-rate of 39.8\%. In addition, we develop a novel method of adapting pre-trained masked language models to generate measurements of how much a generated sentence exemplifies the meaning of a word. When this method is used to rerank a set of 5 sentences generated by LLMs for each word for use in the OxfordEval measurement, the win-rate further increases to 85.1\%. We estimate that one end-to-end run of the generation, reranking, and evaluation steps for a set of 8000 word senses and definitions costs less than \$50. Put together, we hope the work provides a refreshed low-cost baseline for the development of methods to generate high quality dictionary example sentences for the benefit of language learners. 


\section{Related Work}
Recent work has demonstrated that language models can be trained on existing sets of headwords and example sentences from dictionaries and other corpora for automatic example sentence generation. \citet{barba2021exemplification} and \citet{he2022controllable} adapt encoder-decoder architectures trained on reference sentences for each word sense to generate new sentences on unseen headwords, with \citet{he2022controllable} demonstrating the ability to control for length and lexical complexity. \citet{harvill2023one} demonstrate that sentence generation can also be performed in a one-shot manner using just a reference sentence without word sense labels (which are less abundantly available), using an autoencoder trained to generate new sentences conditioned on latent sense representations. 

Earlier studies of dictionary example sentences use a small-scale experimental setup to evaluate the usefulness of dictionary headwords with and without examples on learning outcomes \citep{nesi1996role,frankenberg2014use}. These methods measure the direct impact for learners, but are limited to small-scale experimental settings. Later studies use specific model-based scores to automate the evaluation of sentence quality. \citet{harvill2023one} compared the cosine similarity of pretrained BERT-large word embeddings \cite{scarlini2020more} of the target word w* between generated sentences and example sentences in a golden dataset. \citet{he2022controllable} trained a BERT model to evaluate definition and part-of-speech (POS) accuracy using triplets of (word, definition, example) and (word, POS, example) respectively, with positive examples from Oxford Dicionary and negative examples generated synthetically by replacing words, definitions, POS, and examples. However, these automatic methods evaluate narrow factors (word sense accuracy, POS accuracy and definition accuracy) that have not been shown to be correlated with human-validated preferences for the sentence as a whole.

Recent studies have demonstrated the ability of LLMs to evaluate free-form outputs of LLMs as instruction-following agents \cite{dubois2023alpacafarm} and chatbot assistants \cite{zheng2023judging}, with a high degree of correlation to human preference when performing pairwise comparison on the overall quality of outputs. However, they also warn of position bias, verbosity bias and self-enhancement bias \cite{zheng2023judging}.

\section{Task Definition}
\subsection{Sentence Generation}
The task of dictionary example generation is to generate fluent example sentences for dictionary headwords that effectively illustrate the target word under a specific sense or word meaning provided. Following the convention in \citet{he2022controllable}, the dictionary example sentence task aims to generate an example $E = \{e_1,...,e_T\}$, given a target word $w^*$, a definition $\mathit{D}=\{d_1,..,d_S\}$ and part-of-speech $\mathit{P} \in \{p_1,..,p_K\}$.
\subsection{Dataset}
We use the Oxford Dictionary dataset compiled by \citet{gadetsky2018conditional} and processed by \citet{he2022controllable}. Each entry in this dataset is a single word sense, with its associated lemma, part-of-speech, definition, and example sentences demonstrating the usage of the word in a sentence. Following \citet{he2022controllable}, we note that the dataset contains multiple entries for different inflections of the same lemma (for example, 'allow', 'allowing', 'allows', and 'allowed' for the lemma 'allow' meaning 'let (someone) have or do something'), and keep only the inflection with the highest number of example sentences for each (lemma, POS, definition) triplet. The entire dataset comprises 105,818 Word Senses across the Training, Validation and Test splits. As we do not perform any model training, we focus only on the Validation and Test splits. 
\begin{table}
  \centering
\footnotesize{\bgroup
\def\arraystretch{1.3}%
\begin{tabular}{ccccc}
\toprule
Dataset & Word & Unique & Unique & Average \\
Split & Senses & Words & Lemmas & Number of\\
& & & & Example \\
& & & & Sentences \\
\hline
Validation & 7,931 & 6,992 & 6,311 & 11.0 \\
Test & 7,843 & 6,963 & 6,256 & 11.1 \\
\bottomrule 
  \end{tabular}\egroup}
  \caption{Summary statistics: Number of unique words, word senses, lemmas, and example sentences in our  dataset. Each unique word sense forms the unit of analysis for our study. } 
  \label{tab:dataset-table}
\end{table}
Table \ref{tab:dataset-table} shows the summary statistics for the dataset. The validation set contains 7,931 word senses across 6,992 unique words and 6,311 unique lemmas, with an average of 11.0 example sentences per word sense. The test set contains 7,843 word senses across 6,963 unique words and 6,256 unique lemmas, with an average of 11.1 example sentences per word sense. 
As these splits are derived through random sampling from the Oxford Dictionary dataset, it can give a reasonable measure of the effectiveness of methods when scaled up to the entire dictionary while managing experimentation costs. Also, as we use the same preprocessing steps as 
\citet{he2022controllable}, we can compare our results with their generated sentences. 

Following \citet{he2022controllable}, we also augment the dataset with word occurrence frequencies from the one-billion-word dataset \cite{chelba2013one}, to be used as a proxy for word complexity \cite{paetzold2016semeval} for supplementary analysis. 

\subsection{Sentence Evaluation}

We use LLMs to perform competitive evaluations of generated sentences against example sentences in the Oxford Dictionary dataset. As past studies have already demonstrated the ability to generate good example sentences that are fluent and semantically accurate (\citet{he2022controllable,harvill2023one,barba2021exemplification}), we propose that the next step would be to attempt to competitively identify sentences of the highest quality so that we can maximise the efficiency of language learning for dictionary users. 

We apply the recent approaches in pairwise output ranking \citep{zheng2023judging} and automating evaluations by using LLM as evaluators \citep{dubois2023alpacafarm}, with one change: instead of performing comparisons of two LLM outputs, we use the example sentences within the Oxford dictionary dataset as the baseline comparison. 

Formally, for each sentence generation method $\mathbf{G}: (w^*,D,P) \to E$, we calculate the win rate of the set of generated sentences $\{E_1,...,E_M\}$ over a baseline set of dictionary examples $\{E^*_M,...,E^*_M\}$ using an evaluator $\mathbf{E}: (E,E^*) \to L\in\{0,1\}$, where $L=0$ indicates $E$ is preferred over $E^*$ and $L=1$ indicates the reverse. For each word sense, we use the first\footnote{Our experiments find no difference in quality based on order of sentences in the Oxford Dictionary dataset. Computing the OxfordEval metric for randomly selected sentences against the first sentence in the example sentence set achieved a win-rate of 49.9\%.} sentence amongst the example sentences in the Oxford dataset as the baseline dataset. By averaging over the labels $L$, we calculate the win-rate of a proposed sentence generation method over existing Oxford dictionary examples, which we define as the \textbf{OxfordEval} win-rate. A win-rate of above 50\% will indicate that the generated sentences, on average, would be of higher quality than the sentences in the Oxford dictionary dataset. 


\begin{table}[h]
\begin{tabular}
{p{4.5cm}cc}
\hline  {} & Agreement (\%)\\
\hline
Claude-V1 & \textbf{73.6} \\
Llama-2-70B Chat & 71.6 \\
Claude-V2 & 67.1 \\
Mistral-7B Instruct & 60.0\\
Llama-2-7B Chat & 52.1\\
Llama-2-13B Chat & 47.7\\
\hline
\end{tabular}
\caption{\label{table:llm_evaluators} Agreement rate between LLM evaluators and human-annotated preferences, with highest value indicated in bold.}
\end{table}

To validate our approach, we (the authors) annotated 501 examples of pairwise comparison between Claude-Instant generated sentences and baseline Oxford sentences using the Amazon SageMaker Ground Truth platform. The order of sentences presented was randomly flipped half the time and the source of each sentence cannot be seen by the annotator. For each example, we collected annotations from 2 different annotators and only kept examples with inter-annotator agreement. After doing so, we have 447 samples, with more than 89.2\% inter-annotator agreement. We then compare these results with the outputs of different LLMs used as evaluators, and present the results in Table \ref{table:llm_evaluators}. We choose to use Claude-V1 for our subsequent experiments, as it displayed the highest agreement with human preference and also the highest score on ChatbotArena leaderboard\footnote{https://chat.lmsys.org/} (which measures agreement between LLM outputs and human preference) at the time of study. The prompt used for sentence evaluation is given in Appendix \ref{sec:eval sentences}. 

Beyond the OxfordEval metric, we also track secondary metrics that can be relevant for the selection of preferred dictionary sentences. To measure sentence length, we track the average and standard deviation of the number of words in each sentence. To measure readability, we track the average and standard deviation of the Flesch-Kincard Grade Level (FKGL) \cite{kincaid1975derivation} of each sentence. The averages of these metrics would provide insight on the length and readability of the typical sentence, while the standard deviations would provide insight on the diversity in terms of length and readability across sentences. For these secondary metrics, we propose that they should merely be used to provide an indication of the nature of sentences generated, rather than values that are blindly maximised or minimised, as the requirements could vary depending on the specific needs being addressed. For example, lower average lengths and FKGL scores could be desirable for new language learners, but intermediate learners might benefit from access to more complex and longer sentences. 


\section{Sentence Generation}
\begin{figure*}
  \centering \includegraphics[width=\textwidth,height=\textheight,keepaspectratio]{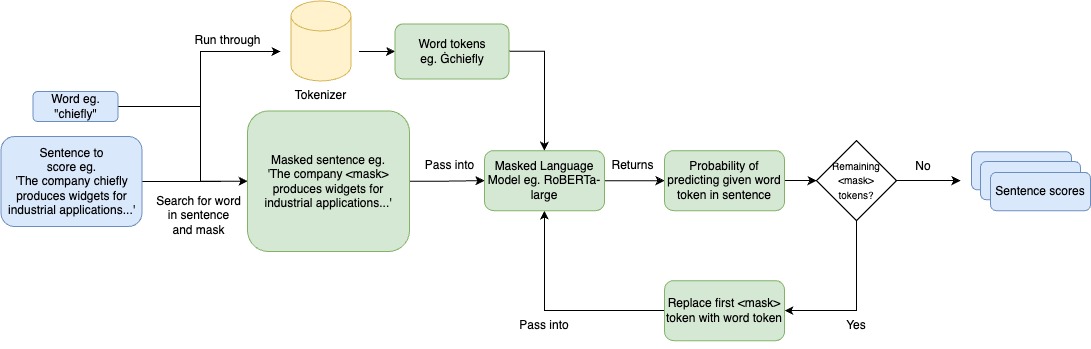}
  \caption{Measuring sentence exemplification with masked language models. First, we search for the word within the sentence and replace them with masked tokens. We then use the masked language model to measure the probability of reconstructing the same sentence in order to get the exemplification score.}
\end{figure*}

\begin{table*}[h]
\begin{center}
\footnotesize{\begin{tabular}
{lp{115mm}c}
\toprule  Source & Sentence & Exemplification \\
{} & {} & Score \\
\hline \\
Oxford & There was a dull pain in his lower jaw. & 0.031 \\
Oxford & All through her tantrum she felt the pain inside of her, but with after a half an hour her pain subsided into a dull ache. & 0.969 \\
Oxford & My finger is recovering well, I'm in no pain from that quarter, although I have a dull ache in my leg where I was shot full of medication. & 0.527 \\
\hline \\
Claude & While the headache had been throbbing and sharp earlier, now it was just a dull ache that was easy to ignore. & 0.973 \\
Claude & I felt a dull ache in my leg from overexerting myself during the hike. & 0.257 \\
Claude & After taking some pain medication, the ache in my back changed from a sharp pain to a dull one. & 0.890 \\
\bottomrule
\end{tabular}}
\caption{\label{table:examples} Exemplification scores for sample sentences taken from Oxford Dictionary and generated by Claude-v1 for 'dull' (adj. (of pain) indistinctly felt; not acute). We use exemplification scores to provide a measurement of how much each sentence exemplifies the meaning of the word. Higher scores indicate better exemplification of meaning.}
\end{center}
\end{table*}

In this section, we present the \textbf{FM-MLM} (\textbf{F}oundational \textbf{M}odel - \textbf{M}asked \textbf{L}anguage \textbf{M}odel) method for generating quality dictionary example sentences at low cost. FM-MLM uses a three-step process. First, foundational LLMs are used to generate a set of candidate sentences. Next, pre-trained masked language models are adapted with a novel method to score how much each candidate sentence exemplifies the meaning of the target word. Finally, the sentence with the highest score is selected to be the final generated sentence for evaluation.

\subsection{Generating sentences with LLMs}
We use zero-shot prompts to generate candidate sentences. The LLM is given the word, POS, and definition, and then asked to generate a sentence that illustrates the given definition of the word (the full prompt is given in Appendix B). We repeat the generation 5 times per word sense to obtain a set of 5 generated sentences. We observe that LLMs can sometimes fail to conform to our specified output format or refuse to generate sentences for certain sensitive words. In these cases, we retry the generation for up to 10 times per word sense. If sentences are not successfully generated after 10 retries, we abandon the process for that word sense and impute a dummy blank sentence for subsequent evaluation. This occurred for 84 out of 7931 word senses in the validation set.  


\subsection{Measuring exemplification using masked language models}

In this section, we propose a novel method to identify sentences that exemplify the meaning of each word, through creatively adapting pre-trained masked language models.

Prior work \cite{barba2021exemplification} has proposed that Word Sense Disambiguation (WSD) models can be used to measure the extent to which a sentence exemplifies the meaning of a word. If a sentence exemplifies the meaning of a word, then a WSD model should be able to identify the correct sense of the word with high probability. This approach, however, has some limitations. Consider for example the case where a word has only a verb and a noun sense, such as pedal, trivial sentences like “I will \emph{pedal} this” (v. move (a bicycle) by working its pedals) and “This is a \emph{pedal}” (n. each of a pair of foot-operated levers used for powering a bicycle or other vehicle propelled by the legs) could almost perfectly disambiguate between word senses, but do little to illustrate the meaning of the word to a language learner. 

We propose that a higher standard be used. Instead of simply disambiguating between the \emph{senses of the same word}, we propose that exemplification can be measured by disambiguating against \emph{all other words}. 

Figure 1 illustrates how LLMs pre-trained on the masked language modelling task can be adapted to measure this. We first mask the target word in the sentence, and then compute the probability that the word masks correspond to the target word. A sentence that exemplifies the meaning of a word will receive a high probability score. We use this probability score as a proxy measure for the degree of exemplification of the word meaning in a sentence, and select the sentence with the highest score to perform evaluation. 

For example, for the word 'poor' (adj. of a low or inferior standard or quality), sentences in the Oxford Dictionary dataset range in exemplification from 'Information dissemination and knowledge of law are poor at this level.' (p=0.015) to 'A local police official blamed shoddy construction and the poor quality of the cement.' (p=0.942). In the former sentence, we can see that the word 'poor' could easily be replaced by many other words, even with antonyms such as 'excellent'. We suggest that such sentences are too general in nature and may not be as effective for language learners.

Table \ref{table:examples} illustrates the exemplification score of various sentences using the word 'dull' (adj. (of pain) indistinctly felt; not acute). The sentence 'There was a dull pain in his lower jaw.' is rated as having a low exemplification score, because we can easily replace 'dull' with antonyms ('There was a sharp pain in his lower jaw.') or unrelated words ('There was a constant pain in his lower jaw."). In contrast, the sentence "While the headache had been throbbing and sharp earlier, now it was just a dull ache that was easy to ignore.' has a higher exemplification score because it contains elements that hint towards the meaning of 'dull', such as its contrast to 'throbbing and sharp' and being 'easy to ignore'. 

\section{Model Development}
\subsection{Development Environment}
We perform our experiments on Amazon Sagemaker. For sentence generation and evaluation, AWS Bedrock APIs are used to access Claude-V1, Claude-Instant-V1 \cite{anthropic2023introducing}, Claude-V2 \cite{anthropic2023claude2}, Llama-2 13B Chat and Llama-2 70B Chat \cite{touvron2023llama} models, and Amazon Sagemaker was used to deploy Llama-2-7B Chat and Mistral-7B-Instruct v0.1 \cite{jiang2023mistral}. The following pre-trained language models were used and run using the HuggingFace transformers library: BERT-base, BERT-large, RoBERTa-base, and RoBERTa-large \cite{kenton2019bert,liu2019roberta}. Experiments were run for a single replication for each model configuration. 

\subsection{Model Configurations and Hyperparameters}

\begin{table}[t]
\begin{tabular}
{p{3.3cm}p{3.3cm}}
\hline  Hyperparameter & Values\\
\hline \\
\emph{Sentence Generation} \\ 
LLM & \textbf{Claude-Instant-V1}, Llama-2-Chat (7B, 13B, 70B), Mistral-7B Instruct\\
Batching Method & \textbf{One at a time}, All-at-once \\
Number of Sentences & 1-5 \textbf{(5)} \\
Inputs & \textbf{POS+Def}, POS Only, Definition Only \\
\\
\emph{Sentence Selection} \\ 
Selection Criteria & \textbf{MLM-Score}, First Sentence \\
MLM Model & RoBERTa-Base, \textbf{RoBERTa-Large}, BERT-Base, BERT-Large\\ \\
\hline
\end{tabular}
\caption{\label{table:hyperparameter_search} Hyperparameter search space. Bolded text indicates hyperparameters used in the FM-MLM model.}
\end{table}

Table \ref{table:hyperparameter_search} shows the search space and final settings for FM-MLM's configurations and hyperparameters. We call Claude-instant-v1 with temperature of 0.9 to generate sentences one at a time to accumulate 5 sentences for each word sense, with a buffer for up to 10 retries in the event of failures. Top P was set to 1.0 and Top K was set to 500. The sentences are scored with a RoBERTa-large model, with the highest scoring sentence selected for evaluation against the example sentence in the Oxford Dictionary dataset.  These hyperparameters were tuned manually to maximise the win-rate on the validation set, with one exception: while we could have further increased the number of sentences generated for each word sense to give more options to select from, we limited the number to 5 to manage computational costs. 

With these hyperparameters, the estimated cost of generating and evaluating sentences is US\$45 for the validation set of 7,931 sentences. This is broken down into \$21.4 for the use of Claude-instant v1 to generate 5 example sentences, \$0.55 to generate exemplification scores with RoBERTa-large over 45 minutes on g4dn.xlarge EC2 instances, and \$23.04 to evaluate the outputs with Claude v1. 

\subsection{Ablation Experiments}
To validate the model configuration and hyperparameters, ablation experiments were conducted where we start with our model and then ablate or vary individual components one at a time, keeping all others constant, and then examine the differences in performance on the validation set. 

For our ablation experiments, we focus on the primary metric (the overall win-rate of the generated sentence for each word sense against the first Oxford Dictionary example sentence in the validation set), and supplement with additional analysis where relevant to understand potential explanations for the trends observed. 

\subsubsection{Choice of LLM for sentence generation}

\begin{table}[h]
\begin{tabular}
{p{4.8cm}cc}
\hline  {} & Win-rate (\%)\\
\hline
FM-MLM (Claude-Instant) & 85.6\\
Llama-2-70B Chat & 82.7\\
Llama-2-13B Chat & 79.8\\
Llama-2-7B Chat & 78.6\\
Mistral-7B Chat & 72.2\\
No LLM (Oxford Sentences) & 55.9\\
\hline
\end{tabular}
\caption{\label{table:llm_generator} Performance when the LLM used for sentence generation is varied on the validation set.}
\end{table}

Table \ref{table:llm_generator} shows how the win-rate varies when various open-source and closed-source LLMs are used for sentence generation. We see that the win-rate is highly sensitive to the choice of LLM for sentence generation, with win-rates ranging from 72.2\% (Mistral-7B-chat) to 85.6\% (Claude-instant-v1) depending on the LLM used. 

Also, we report the win-rate when the LLM generation step is ablated, and sentences are selected based on the MLM exemplification metric on the full list of example sentences in the Oxford Dictionary dataset. In this setting, the win-rate is 55.9\%, demonstrating the ability of the MLM exemplification metric to pick higher quality sentences among Oxford Dictionary examples. 

\subsubsection{Batched Generation vs Repeated Single Sentence Generation}

\begin{table}[h]
\begin{tabular}
{p{4.8cm}cc}
\hline  {} & Win-rate (\%)\\
\hline
FM-MLM (One-by-one) & 85.6\\
Batched Generation (5 at once) & 78.5\\
\hline
\end{tabular}
\caption{\label{table:generation_batch} Performance when the batching strategy for sentence generation is varied on the validation set.}
\end{table}

One interesting finding observed was that FM-MLM was most effective when the LLMs were prompted to generate only a single sentence at a time (with the process repeated multiple times to get the final set of sentences to select from). As we can see in Table \ref{table:generation_batch}, changing the prompt to perform batched generations (“Generate 5 sentences…” instead of “Generate a sentence”) resulted in the win-rate dropping from 85.6\% to 78.5\%. 

We hypothesize that this could be because the LLM might be primed to consider factors beyond simply maximising the quality of each individual sentence, such as ensuring diversity in the outputs. Two observations support this point: (1) taking the first sentence generated by the single sentence generation approach (win-rate = 84.4\%) outperformed taking the first sentence generated by the batched generation approach (win-rate = 81.3\%); (2) the win-rate for sentences decline as we go from the 1st sentence (81.3\%) to the 5th sentence (Win-rate for 1st sentence = 81.3\%; 2nd = 77.5\%; 3rd = 76.4\%; 4th = 74.8\%; 5th = 75.6\%). Using the MLM scores to select sentences (78.5\%) also gives a lower score than simply picking the 1st sentence in this case, possibly because the MLM scores only measure exemplification and not other aspects of sentence quality. 

\subsubsection{Inclusion / Exclusion of Definition and POS}

\begin{table}[h]
\begin{tabular}
{p{4.8cm}cc}
\hline  {} & Win-rate (\%)\\
\hline
FM-MLM (Definition + POS) & 85.6\\
Ablate POS (Definition only) & 82.7\\
Ablate Definition (POS only) & 55.6\\
\hline
\end{tabular}
\caption{\label{table:def_pos} Performance when the inputs for sentence generation are varied on the validation set.}
\end{table}

Table \ref{table:def_pos} shows how the win-rate varies when the definition and POS of the word were removed from the generation prompt. We observe a marginal drop in win-rate when POS is removed, and a very substantial drop in win-rate when word definitions are removed. 

There are two possible hypotheses for this. (1) the LLMs rely heavily on the given definition in the prompt to generate effective sentences, rather than latent knowledge of word definitions from pretraining. (2) for polysemous words, there is confusion over which definition to generate the sentence for. Further analysis suggests that hypothesis (1) might play a heavier role. When we compute the win-rate for words with a single sense (n = 5635; win-rate = 60.4\%), it is slightly higher than polysemous words (n = 2296; win-rate = 43.3\%), but not to the extent that it can explain the huge drop in win-rate. 

\subsubsection{Choice of MLM model to score sentences}

\begin{table}[h]
\begin{tabular}
{p{4.8cm}cc}
\hline  {} & Win-rate (\%)\\
\hline
FM-MLM (RoBERTa-Large) & 85.6\\
RoBERTa-Base & 85.4\\
BERT-Large & 85.5\\
BERT-Base & 85.2\\
Ablate Selection & 84.4\\
\hline
\end{tabular}
\caption{\label{table:mlm_picker} Performance when the sentence selection strategy is varied on the validation set.}
\end{table}

Table \ref{table:mlm_picker} shows the results when the MLM model used to measure exemplification in generated sentences is varied. The general trend observed is that bigger and more advanced models achieve better performance, but the effect is only marginal. 

We also considered the possibility that MLM models might be more effective in measuring exemplification for words that are represented as a single token, as past work has suggested that it is more challenging to model words that are fragmented into multiple tokens \cite{domainroberta}. When the validation set is split into single-token words (n=4090) and multi-token words (n=3641), we conversely observe a lower win rate for single-token words (84.0\%) as compared to multi-token words (87.3\%). However, we hesitate to draw a general conclusion that the MLM model is more effective in scoring multi-token words, as there could be confounding factors such as LLMs being less reliable for multi-token words which are typically of higher difficulty in nature. 

When no MLM model is used to score sentences (and the first sentence is picked), the win-rate drops from 85.6\% to 84.4\%. 

\subsubsection{Choice of LLM evaluator model}

\begin{table}[h]
\begin{tabular}
{p{4.8cm}cc}
\hline  {} & Win-rate (\%)\\
\hline
Claude-V1 as Evaluator & 85.6 \\
Llama2-70B Chat as Evaluator & 86.9 \\
\hline
\end{tabular}
\caption{\label{table:llm_evaluator_ablation} Performance when switching the LLM evaluator between the top two evaluators aligned with human preference. The evaluated model is FM-MLM (Claude-Instant) over the baseline Oxford dictionary sentences.}
\end{table}

Table \ref{table:llm_evaluator_ablation} shows how the win-rate over the baseline Oxford sentences changes when we switch the evaluator model from Claude-V1 to Llama-2-70B Chat. These two evaluator models were found to have the two highest alignment to human preference as shown in Table \ref{table:llm_evaluators}. Our findings show that the win-rates we observe for FM-MLM are robust across LLM evaluators. Our findings are also in alignment with \citet{zheng2023judging} who found that LLMs demonstrate no preference for LLMs of the same family, and are only biased if the same LLM is used to evaluate outputs generated by itself.

\section{Main Results}

\begin{table*}[h]
\begin{center}
\footnotesize{\begin{tabular}
{lccccc}
\toprule  {} & Win-Rate (\%) & Words (Avg) & Words (SD)  & FKGL (Avg) & FKGL (SD) \\
\hline \\
$[$1] FM-MLM & \textbf{85.1} & 16.9 & 4.0 & 8.9 & 3.2 \\
$[$2] FM Only & 83.9 & 16.5 & 3.9 & 8.8 & 3.2 \\
$[$3] Oxford (+MLM Selection) & 56.7 & 19.9 & 7.1 & 9.5 & 3.8 \\
(Baseline) Oxford (1st Sentence)  & (50.0) & 18.7 & 7.4 & 9.1 & 3.9 \\
$[$4] CDEG-Beam \cite{he2022controllable} & 39.8 & 11.2 & 1.3 & 5.3 & 2.9 \\
$[$5] CDEG-Greedy \cite{he2022controllable} & 39.3 & 11.7 & 1.1 & 5.7 & 2.9 \\\\
\bottomrule
\end{tabular}}
\caption{\label{table:main_results} Main results on the test set for various approaches to dictionary example sentence generation. Models are sorted by the OxfordEval win-rate (bold indicates the highest win-rate). The average and standard deviation of the number of words and Flesch-Kincaid Grade-Levels provide indications of the length and readability of sentences.}
\end{center}
\end{table*}

Table \ref{table:main_results} shows the results of our models on the test set. We present the following models. Model (1) is the full FM-MLM model, Model (2) uses only the 1st sentence generated by the LLM, to evaluate the independent performance of LLMs and the additional contribution of the MLM exemplification scores. Model (3) uses the MLM exemplification scores to select amongst example sentences in the Oxford Dictionary dataset. Models (4) and (5) are the CDEG-Beam and CDEG-Greedy models, evaluated using the outputs provided by \citet{he2022controllable}.  

Sentences generated by the FM-MLM model achieve a win-rate of 85.1\% when compared to example sentences in the Oxford Dictionary dataset, demonstrating the ability of FM-MLM to generate sentences that compete favorably against expert references. Without the use of MLM exemplification scores to select sentences, the win-rate drops slightly to 83.9\%. Using MLM exemplification scores to select amongst samples in the Oxford Dictionary dataset gives a win-rate of 56.7\%. These results demonstrate that the MLM exemplification scores can be used to identify sentences of higher quality from a set. CDEG-Beam and CDEG-Greedy sentences achieve a win-rate of 39.8\% and 39.1\% respectively. This is aligned with results reported in \cite{harvill2023one}, where S2S \cite{harvill2023one} and ExMaker \cite{barba2021exemplification} were found to have slightly lower human-annotated scores, in terms of semantic match and fluency, when compared against gold samples in the Oxford Dictionary and SemCor dataset. We note that the win rate for Claude Instant (batched generation) over baseline Oxford sentences based on 447 agreed human annotations was 76.5\%, which is similar to the win-rate of 78.5\% evaluated by the chosen LLM evaluator shown in Table \ref{table:generation_batch}.

By examining the secondary metrics, we see that sentences generated by FM-MLM has an average (SD) of 16.9 (4.0) words per sentence and an average (SD) FKGL of 8.9 (3.2). These are lower than the sentences in the Oxford Dictionary dataset, which has an average (SD) of 18.7 (7.4) words and an average (SD) FKGL of 9.1 (3.9), indicating that the Oxford Dictionary dataset has a more diverse range of sentences that are also typically more advanced in nature. Sentences generated by CDEG models tend to be shorter, simpler and more diverse in nature, with an average (SD) of 11.2 (1.3) words and average (SD) FKGL of 5.3 (2.9) for CDEG-Beam. 

We also obverse that the MLM selection step tends to favor slightly longer sentences, with the average words and FKGL levels increasing slightly for FM-MLM as compared to FM Only, and Oxford (+MLM Selection) compared to Oxford (1st Sentence). 

\section{Conclusion}

Rapid advancements in LLM research has led to the creation of powerful, general-purpose models that can be flexibly adapted for myriad tasks without the need for expensive customisation efforts. While previous works years before this paper had to rely on compute-intensive model training steps and scarce datasets of word-sense annotations, we are now able to study dictionary example sentence generation at a fraction of the cost. 

In this paper, we demonstrated the FM-MLM model, which uses foundational LLMs to generate dictionary example sentences, and then uses a novel adaptation of pretrained masked language models to provide a measure of how much they exemplify the meaning of words. With automatic evaluation methods of the overall sentence quality that are validated with human preference data (in the form of the OxfordEval metric), we are able to easily perform extensive ablation studies on how various modelling choices and hyperparameters affect the effectiveness of the model, which we note was difficult for past studies. Sentences generated by the FM-MLM model achieve a win-rate of 85.1\% when evaluated competitively against typical sentences in the Oxford Dictionary dataset. 

The high win-rate indicates the potential of adapting the FM-MLM model in practice to generate example sentences for actual use by language learners. However, for safety reasons, we recommend that any real-world application should be carefully supervised and governed to manage risks and limitations which we discuss in the limitations section. 

We identify many potential areas for impactful future work. To improve sentence quality further, researchers can consider methods to customise and tune the foundational models for this specialised task. To work towards real-world use, we can also perform more rigourous and holistic evaluations on generated sentences for accuracy, fluency, and other factors of interest. Evaluation can also be conducted on sets of \emph{multiple sentences}, rather than single sentences as we have done in this study, in order to better capture factors such as sentence diversity that would apply for a real dictionary. Lastly, the work can also be extended for other languages. We hope that our work enables these studies by providing a refreshed low-cost baseline reference for good quality sentence generation and evaluation that future researchers can build on. 

\section*{Limitations}

While FM-MLM achieves an impressive win-rate over real sentences in the Oxford Dictionary dataset, we caution against applying FM-MLM in real-world settings without further study and risk mitigation measures. Notably, the pairwise evaluation method we use only provides a relative measure of quality, and there is a risk of endorsing erroneous sentences when both candidate sentences are of poor quality. The accuracy of evaluation is also dependent on the capabilities of LLMs, and while we demonstrate agreement of LLM evaluators with human preferences at a level comparable to \citet{dubois2023alpacafarm}, the agreement rate is still lower than the agreement among human annotators. Possible ways to manage risks would include having experts to vet generate sentences before real-world use. 

We also note that dictionary example sentences could have purposes beyond supporting language learners, such as documenting the etymology of words and language use over time as is the case for the Oxford English Dictionary\footnote{https://www.oed.com/}. Our work focuses only on human preferences of language learners, and does not aim to exhaustively address all the possible purposes of dictionary example sentences. 

We acknowledge the potential for biases of using LLMs as evaluators. We mitigate this potential limitation by studying the alignment with human preference, and also with ablating the choice of LLM evaluator models for our specific task. We also acknowledge the potential for test set sentence leakage in the training corpora used to train the generator LLMs. This limitation is mitigated by our evaluation strategy of comparing win-rate over test set Oxford sentences, which mean that generator LLMs would still have to generate sentences preferred to potentially leaked test set sentences.

Lastly, we acknowledge that the human preferences dataset, which was annotated by ourselves, would only represent the preferences of our narrow demographic group and could also contain unseen biases. The representativeness of the dataset can be further improved using a more diverse pool of annotators in future.

\section*{Ethics Statement}
No conflicts of interest are declared by the authors. The dataset does not contain any personally identifiable data.


\section*{Acknowledgements}
We would like to thank reviewers and meta-reviewers for their helpful comments. We would also like to thank Evelyn Ng, Sheldon Liu, Eric Ho, Annalyn Ng and Guang Yang for support that made this paper possible. 
\bibliography{anthology,custom}

\begin{thebibliography}{29}
\expandafter\ifx\csname natexlab\endcsname\relax\def\natexlab#1{#1}\fi

\bibitem[{{Anthropic}(2023{\natexlab{a}})}]{anthropic2023claude2}
{Anthropic}. 2023{\natexlab{a}}.
\newblock Claude 2.
\newblock \url{https://www.anthropic.com/news/claude-2}.
\newblock Accessed: 2024-02-01.

\bibitem[{{Anthropic}(2023{\natexlab{b}})}]{anthropic2023introducing}
{Anthropic}. 2023{\natexlab{b}}.
\newblock Introducing claude.
\newblock \url{https://www.anthropic.com/news/introducing-claude}.
\newblock Accessed: 2024-02-01.

\bibitem[{Barba et~al.(2021)Barba, Procopio, Lacerra, Pasini, Navigli et~al.}]{barba2021exemplification}
Edoardo Barba, Luigi Procopio, Caterina Lacerra, Tommaso Pasini, Roberto Navigli, et~al. 2021.
\newblock Exemplification modeling: Can you give me an example, please?
\newblock In \emph{IJCAI}, pages 3779--3785.

\bibitem[{Bommasani et~al.(2021)Bommasani, Hudson, Adeli, Altman, Arora, von Arx, Bernstein, Bohg, Bosselut, Brunskill et~al.}]{bommasani2021opportunities}
Rishi Bommasani, Drew~A Hudson, Ehsan Adeli, Russ Altman, Simran Arora, Sydney von Arx, Michael~S Bernstein, Jeannette Bohg, Antoine Bosselut, Emma Brunskill, et~al. 2021.
\newblock On the opportunities and risks of foundation models.
\newblock \emph{arXiv preprint arXiv:2108.07258}.

\bibitem[{Chelba et~al.(2013)Chelba, Mikolov, Schuster, Ge, Brants, Koehn, and Robinson}]{chelba2013one}
Ciprian Chelba, Tomas Mikolov, Mike Schuster, Qi~Ge, Thorsten Brants, Phillipp Koehn, and Tony Robinson. 2013.
\newblock One billion word benchmark for measuring progress in statistical language modeling.
\newblock \emph{arXiv preprint arXiv:1312.3005}.

\bibitem[{Dubois et~al.(2023)Dubois, Li, Taori, Zhang, Gulrajani, Ba, Guestrin, Liang, and Hashimoto}]{dubois2023alpacafarm}
Yann Dubois, Xuechen Li, Rohan Taori, Tianyi Zhang, Ishaan Gulrajani, Jimmy Ba, Carlos Guestrin, Percy Liang, and Tatsunori~B Hashimoto. 2023.
\newblock Alpacafarm: A simulation framework for methods that learn from human feedback.
\newblock \emph{arXiv preprint arXiv:2305.14387}.

\bibitem[{Frankenberg-Garcia(2014)}]{frankenberg2014use}
Ana Frankenberg-Garcia. 2014.
\newblock The use of corpus examples for language comprehension and production.
\newblock \emph{ReCALL}, 26(2):128--146.

\bibitem[{Frankenberg-Garcia et~al.(2021)Frankenberg-Garcia, Rees, and Lew}]{frankenberg2021slipping}
Ana Frankenberg-Garcia, Geraint~Paul Rees, and Robert Lew. 2021.
\newblock Slipping through the cracks in e-lexicography.
\newblock \emph{International Journal of Lexicography}, 34(2):206--234.

\bibitem[{Gadetsky et~al.(2018)Gadetsky, Yakubovskiy, and Vetrov}]{gadetsky2018conditional}
Artyom Gadetsky, Ilya Yakubovskiy, and Dmitry Vetrov. 2018.
\newblock Conditional generators of words definitions.
\newblock In \emph{Proceedings of the 56th Annual Meeting of the Association for Computational Linguistics (Volume 2: Short Papers)}, pages 266--271.

\bibitem[{Hanks(2009)}]{hanks2009impact}
Patrick Hanks. 2009.
\newblock The impact of corpora on dictionaries.
\newblock \emph{Contemporary corpus linguistics}, pages 214--236.

\bibitem[{Harvill et~al.(2023)Harvill, Hasegawa-Johnson, Yoon, Yoo, and Yoon}]{harvill2023one}
John Harvill, Mark Hasegawa-Johnson, Hee~Suk Yoon, Chang~D Yoo, and Eunseop Yoon. 2023.
\newblock One-shot exemplification modeling via latent sense representations.
\newblock In \emph{Proceedings of the 8th Workshop on Representation Learning for NLP (RepL4NLP 2023)}, pages 303--314.

\bibitem[{He and Yiu(2022)}]{he2022controllable}
Xingwei He and Siu~Ming Yiu. 2022.
\newblock Controllable dictionary example generation: Generating example sentences for specific targeted audiences.
\newblock In \emph{Proceedings of the 60th Annual Meeting of the Association for Computational Linguistics (Volume 1: Long Papers)}, pages 610--627.

\bibitem[{Jiang et~al.(2023)Jiang, Sablayrolles, Mensch, Bamford, Chaplot, Casas, Bressand, Lengyel, Lample, Saulnier et~al.}]{jiang2023mistral}
Albert~Q Jiang, Alexandre Sablayrolles, Arthur Mensch, Chris Bamford, Devendra~Singh Chaplot, Diego de~las Casas, Florian Bressand, Gianna Lengyel, Guillaume Lample, Lucile Saulnier, et~al. 2023.
\newblock Mistral 7b.
\newblock \emph{arXiv preprint arXiv:2310.06825}.

\bibitem[{Kenton and Toutanova(2019)}]{kenton2019bert}
Jacob Devlin Ming-Wei~Chang Kenton and Lee~Kristina Toutanova. 2019.
\newblock Bert: Pre-training of deep bidirectional transformers for language understanding.
\newblock In \emph{Proceedings of NAACL-HLT}, pages 4171--4186.

\bibitem[{Kilgarriff et~al.(2008)Kilgarriff, Hus{\'a}k, McAdam, Rundell, and Rychl{\`y}}]{kilgarriff2008gdex}
Adam Kilgarriff, Milos Hus{\'a}k, Katy McAdam, Michael Rundell, and Pavel Rychl{\`y}. 2008.
\newblock Gdex: Automatically finding good dictionary examples in a corpus.
\newblock In \emph{Proceedings of the XIII EURALEX international congress}, volume~1, pages 425--432. Universitat Pompeu Fabra Barcelona.

\bibitem[{Kincaid et~al.(1975)Kincaid, Fishburne~Jr, Rogers, and Chissom}]{kincaid1975derivation}
J~Peter Kincaid, Robert~P Fishburne~Jr, Richard~L Rogers, and Brad~S Chissom. 1975.
\newblock Derivation of new readability formulas (automated readability index, fog count and flesch reading ease formula) for navy enlisted personnel.

\bibitem[{Lewis et~al.(2020)Lewis, Ott, Du, and Stoyanov}]{domainroberta}
Patrick Lewis, Myle Ott, Jingfei Du, and Veselin Stoyanov. 2020.
\newblock Pretrained language models for biomedical and clinical tasks: understanding and extending the state-of-the-art.
\newblock In \emph{Proceedings of the 3rd Clinical Natural Language Processing Workshop}, pages 146--157.

\bibitem[{Liang et~al.(2022)Liang, Bommasani, Lee, Tsipras, Soylu, Yasunaga, Zhang, Narayanan, Wu, Kumar et~al.}]{liang2022holistic}
Percy Liang, Rishi Bommasani, Tony Lee, Dimitris Tsipras, Dilara Soylu, Michihiro Yasunaga, Yian Zhang, Deepak Narayanan, Yuhuai Wu, Ananya Kumar, et~al. 2022.
\newblock Holistic evaluation of language models.
\newblock \emph{arXiv preprint arXiv:2211.09110}.

\bibitem[{Liu et~al.(2019)Liu, Ott, Goyal, Du, Joshi, Chen, Levy, Lewis, Zettlemoyer, and Stoyanov}]{liu2019roberta}
Yinhan Liu, Myle Ott, Naman Goyal, Jingfei Du, Mandar Joshi, Danqi Chen, Omer Levy, Mike Lewis, Luke Zettlemoyer, and Veselin Stoyanov. 2019.
\newblock Roberta: A robustly optimized bert pretraining approach.
\newblock \emph{arXiv preprint arXiv:1907.11692}.

\bibitem[{Mishra et~al.(2022)Mishra, Khashabi, Baral, and Hajishirzi}]{mishra2022cross}
Swaroop Mishra, Daniel Khashabi, Chitta Baral, and Hannaneh Hajishirzi. 2022.
\newblock Cross-task generalization via natural language crowdsourcing instructions.
\newblock In \emph{60th Annual Meeting of the Association for Computational Linguistics, ACL 2022}, pages 3470--3487. Association for Computational Linguistics (ACL).

\bibitem[{Nesi(1996)}]{nesi1996role}
Hilary Nesi. 1996.
\newblock The role of illustrative examples in productive dictionary use.
\newblock \emph{Dictionaries: Journal of the Dictionary Society of North America}, 17(1):198--206.

\bibitem[{Paetzold and Specia(2016)}]{paetzold2016semeval}
Gustavo Paetzold and Lucia Specia. 2016.
\newblock Semeval 2016 task 11: Complex word identification.
\newblock In \emph{Proceedings of the 10th International Workshop on Semantic Evaluation (SemEval-2016)}, pages 560--569.

\bibitem[{Penedo et~al.(2023)Penedo, Malartic, Hesslow, Cojocaru, Cappelli, Alobeidli, Pannier, Almazrouei, and Launay}]{penedo2023refinedweb}
Guilherme Penedo, Quentin Malartic, Daniel Hesslow, Ruxandra Cojocaru, Alessandro Cappelli, Hamza Alobeidli, Baptiste Pannier, Ebtesam Almazrouei, and Julien Launay. 2023.
\newblock The refinedweb dataset for falcon llm: outperforming curated corpora with web data, and web data only.
\newblock \emph{arXiv preprint arXiv:2306.01116}.

\bibitem[{Radford et~al.(2019)Radford, Wu, Child, Luan, Amodei, Sutskever et~al.}]{radford2019language}
Alec Radford, Jeffrey Wu, Rewon Child, David Luan, Dario Amodei, Ilya Sutskever, et~al. 2019.
\newblock Language models are unsupervised multitask learners.
\newblock \emph{OpenAI blog}, 1(8):9.

\bibitem[{Scarlini et~al.(2020)Scarlini, Pasini, Navigli et~al.}]{scarlini2020more}
Bianca Scarlini, Tommaso Pasini, Roberto Navigli, et~al. 2020.
\newblock With more contexts comes better performance: Contextualized sense embeddings for all-round word sense disambiguation.
\newblock In \emph{Proceedings of the 2020 Conference on Empirical Methods in Natural Language Processing (EMNLP)}, pages 3528--3539. The Association for Computational Linguistics.

\bibitem[{Stevenson(2010)}]{stevenson2010oxford}
Angus Stevenson. 2010.
\newblock \emph{Oxford dictionary of English}.
\newblock Oxford University Press, USA.

\bibitem[{Touvron et~al.(2023)Touvron, Martin, Stone, Albert, Almahairi, Babaei, Bashlykov, Batra, Bhargava, Bhosale et~al.}]{touvron2023llama}
Hugo Touvron, Louis Martin, Kevin Stone, Peter Albert, Amjad Almahairi, Yasmine Babaei, Nikolay Bashlykov, Soumya Batra, Prajjwal Bhargava, Shruti Bhosale, et~al. 2023.
\newblock Llama 2: Open foundation and fine-tuned chat models.
\newblock \emph{arXiv preprint arXiv:2307.09288}.

\bibitem[{Wang et~al.(2022)Wang, Mishra, Alipoormolabashi, Kordi, Mirzaei, Naik, Ashok, Dhanasekaran, Arunkumar, Stap et~al.}]{wang2022super}
Yizhong Wang, Swaroop Mishra, Pegah Alipoormolabashi, Yeganeh Kordi, Amirreza Mirzaei, Atharva Naik, Arjun Ashok, Arut~Selvan Dhanasekaran, Anjana Arunkumar, David Stap, et~al. 2022.
\newblock Super-naturalinstructions: Generalization via declarative instructions on 1600+ nlp tasks.
\newblock In \emph{Proceedings of the 2022 Conference on Empirical Methods in Natural Language Processing}, pages 5085--5109.

\bibitem[{Zheng et~al.(2023)Zheng, Chiang, Sheng, Zhuang, Wu, Zhuang, Lin, Li, Li, Xing et~al.}]{zheng2023judging}
Lianmin Zheng, Wei-Lin Chiang, Ying Sheng, Siyuan Zhuang, Zhanghao Wu, Yonghao Zhuang, Zi~Lin, Zhuohan Li, Dacheng Li, Eric Xing, et~al. 2023.
\newblock Judging llm-as-a-judge with mt-bench and chatbot arena.
\newblock \emph{arXiv preprint arXiv:2306.05685}.

\end{thebibliography}

\newpage

\appendix
\setcounter{footnote}{4}
\section{Prompt for Evaluating Sentences \protect \footnotemark }
\fbox{
    \label{sec:eval sentences}
    \begin{minipage}{0.4\textwidth}
        Select the output (a) or (b) that best matches the given instruction. Choose your preferred output, which can be subjective. Your answer should ONLY contain: Output (a) or Output (b). Here's an example: \\ \\
        Example Task: \\ 
        Instruction: Give a description of the following job: "ophthalmologist" \\ \\
        \#\# Output (a): An ophthalmologist is a medical doctor who specializes in the diagnosis and treatment of eye diseases and conditions. \\ 
        \#\# Output (b): An ophthalmologist is a medical doctor who pokes and prods at your eyes while asking you to read letters from a chart. \\ \\ 
        \#\# Which is best, Output (a) or Output (b)? \\ 
        Answer: Output (a) \\ 
        Explanation: Here the answer is Output (a) because it provides a comprehensive and accurate description of the job of an ophthalmologist. In contrast, output (b) is more of a joke. \\ \\ 
        Now is the real task, do not explain your answer, just say Output (a) or Output (b). \\ \\ 
        Actual Task: \\
        Instruction: \\ 
        Construct a sentence with the word \{word\}.\\ 
        The sentence must illustrate the definition of the word given here: \{definition\}. \\ The part-of-speech of the word within the context of the sentence should be a \{POS\}.\\ \\ \#\# Output (a): \{Oxford Dictionary 1st Sentence\} \\ \#\# Output (b): \{Candidate Sentence\} \\ \\ \#\# Which is best, Output (a) or Output (b)?
    \end{minipage}
}
\setcounter{footnote}{4}
\section{Prompt for Generating Sentences\protect \footnote{The prompt highlights the main content and structure of the prompt without the LLM-specific chat template format. LLM-specific chat template is followed during generation with LLMs.}}
\fbox{
    \label{sec:generate sentences}
    \begin{minipage}{0.4\textwidth}

        Using the definition and part-of-speech provided in tags, construct one sentence that illustrates the definition of the given word within the sentence provided. Go straight to the answer with no introduction. \\ \\
        Here are examples: \\
        Using the definition and part-of-speech provided in tags, construct one sentence that illustrates the definition of the word "airstrip" within the sentence provided. Go straight to the answer with no introduction. \\
        <definition> \\
        a strip of ground set aside for the take-off and landing of aircraft. \\
        </definition> \\
        <part-of-speech> \\
        Noun \\
        </part-of-speech> \\
        Assistant: \\
        <sentence>The site has its own airstrip and light aircraft service, and its own small marina. \\
        </sentence> \\
        \{ examples \} \\ \\ \\
        Human: \\
        Using the definition and part-of-speech provided in tags, construct one sentence that illustrates the definition of the word "\{word\}" within the sentence provided. Go straight to the answer with no introduction. \\
        <definition> \\
        \{ definition \} \\
        </definition> \\
        <part-of-speech> \\
        \{part-of-speech\} \\
        </part-of-speech>
    \end{minipage}
}

\end{document}